\documentclass[sigconf]{acmart}

\usepackage{booktabs} 
\usepackage{cuted}
\usepackage{float}
\usepackage{graphicx}

\usepackage{array}
\newcolumntype{L}[1]{>{\raggedright\let\newline\\\arraybackslash\hspace{0pt}}m{#1}}
\newcolumntype{C}[1]{>{\centering\let\newline\\\arraybackslash\hspace{0pt}}m{#1}}
\newcolumntype{R}[1]{>{\raggedleft\let\newline\\\arraybackslash\hspace{0pt}}m{#1}}

\copyrightyear{2018} 
\acmYear{2018}
\setcopyright{rightsretained}
\acmConference[KDD]{Paper for MINING AND LEARNING FROM TIME SERIES WORKSHOP}{2018}{London, UK}

\begin{document}
\title[Integrating Deep Learning and Markov Transition Field for Fraud Detection]{Sequential Behavioral Data Processing Using Deep Learning and the Markov Transition Field in Online Fraud Detection}
\author{Ruinan Zhang}
\affiliation{%
  \institution{CreditX Inc.}
  \city{Shanghai}
  \state{China}
}
\email{zhangrn@creditx.com}

\author{Fanglan Zheng}
\affiliation{%
  \institution{JD Finance Inc.}
  \city{Beijing}
  \state{China}
}
\email{zhengfanglan1@jd.com}

\author{Wei Min}
\orcid{1234-5678-9012}
\affiliation{%
  \institution{CreditX Inc.}
  \city{Shanghai}
  \state{China}
}
\email{minw@creditx.com}

\setlength{\parindent}{0pt}
\settopmatter{printacmref=false}

\begin{abstract}
Due to the popularity of the Internet and smart mobile devices, more and more financial transactions and activities have been digitalized. Compared to traditional financial fraud detection strategies using credit-related features, customers are generating a large amount of unstructured behavioral data every second. In this paper, we propose an Recurrent Neural Netword (RNN) based deep-learning structure integrated with Markov Transition Field (MTF) for predicting online fraud behaviors using customer's interactions with websites or smart-phone apps as a series of states. In practice, we tested and proved that the proposed network structure for processing sequential behavioral data could significantly boost fraud predictive ability comparing with the multilayer perceptron network and distance based classifier with Dynamic Time Warping(DTW) as distance metric.
\end{abstract}

\maketitle


\section*{Introduction}

After the boom of big data beginning in 2010 \cite{Fujitsu}, a significant amount of sequential data have been collected in domains such as finance and bioinformatics. Methodologies on how to mine information from sequences have drawn considerable attentions in both the academia and industry. Particularly, multivariate sequence classification has become very important in a wide range of real-world applications \cite{zheng}. Traditional approaches, such as word2vec, transforming sequential data into some kind of  feature vectors spaces \cite{Xing2010} have been proven effective but still limited in scenarios, especially for multivariate series where each series element contains information across different domains.
\bigskip

Since 2006, the techniques developed from deep neural networks have made many breakthroughs in some domains of sequential data processing, such as language processing and speech recognition \cite{Bengio2009}. One successful deep-learning architecture used in the field of multidimensional data processing is convolution neural networks (CNN) \cite{Lecun95}. One-dimensional CNNs exploit transitional invariance between elements within sequences by extracting features through receptive fields \cite{HubelandWiesel1962} and learning with weight sharing, becoming a state-of-the-art approach in various sequence-processing applications, such as speech recognition and computer vision tasks \cite{Krizhevsky2012}. 
\bigskip

To counter the challenge of multivariate nature of sequential data, MC-DCNN \cite{zheng} (Multi-Channel Deep Convolutional Network) solve the problem by decomposing it into univariate sequences and combining the results of the feature mapping processes of CNN. Techniques in computer vision can also be transfered by encoding time series by applying GASF (Gramian Angular Summation Field) \cite{Oates2015} as a visual approach of extracting information from a series. 
\bigskip

As a co-effort of data scientists from e-commerce and financial industry, we introduce a task-specifical designed and engineered deep-learning network structure against online fraud attacks. Our goal is to overcome the challenges of using state-streaming data in financial scenarios like streaming user behavior data to build effective predictive models with innovative network design.
\bigskip

\subsection*{Problem Background}

Contributed by the prevalence of fast Internet service, smartphones, and advanced online payment platforms, more financial activities happen in the form of non-cash payments on any kinds of online platforms. In the US, non-cash payments have contributed over 144 billion payments with a value of almost 178 trillion dollars in 2015, up nearly 21 billion payments and 17 trillion dollars in value from 2012 \cite{FDR study}. According to a Nilson report \cite{Nilson}, fraudulent activities cost about 11.2 billion dollars worldwide in 2012, and the number has increased almost by 100$\%$ up to 21.84 billion dollars.  The prevalence of online services and mobile devices also has exposed business like online shopping sites and financial services agencies with online platforms to fraudulent attacks. In the online lending market, without an effective fraud detecting strategy, some financial agencies suffer fraud rates over 20\% on their loan services. Financial institutes(typically banks) defend against fraud attacks by implementing chips in credit cards and using algorithms such as experts-defined rules, logistic regression, and support vector machine; However, traditional approaches have been shown to be inapplicable to online shopping and financial services business due to the following challenges:

\subsubsection*{Complexity of Data} 

Online transactions usually happen on websites and smart mobile devices. The data that can be used to evaluate risks are incredibly diverse. One of the most challenging tasks during the stage of data preprocessing is to format data from different types of devices so that data can be evaluated fairly across platforms. Moreover, the behavioral data could cover many fields associated with the object , which include actions, device info, locations, and page content. The challenge relies on how to encode and represent information from different domains into a single state so that information on inter-state relationships can be maintained instead of the traditional approach of extracting aggregated statistical features on every domain. 

\subsubsection*{Volume of Data}

Under the online shopping scenario, data on millions of sessions containing information on IP addresses, devices, URLs, timing, and so on are collected. In Fanglan's experiment, every session is populated with 200 action states, and each action states contains information on different domains can be encoded into over 5000 features. Traditional models such as logistic regressions or ensembling tree models could suffer from a lack of degree of freedom or overfitting of  the dataset with such high dimensionality. For example, for the data collected in the e-commerce , if one flatten all the features on every state, 5204 features on 200 states would result in more than 1,000,000 features. Such high dimensionality combined with a large amount of records makes even modern computational servers hard to process from data storage and computationally algorithmic perspectives.

\subsection*{Description of Data}

Taking finance loans as an example, users behaviors and transitions between different pages before submission of a loan application are recorded on the hosted finance smart-phone app or loan application websites. Users' interactions with the platform are viewed as series of discrete events. At every event, the type of a page, time duration, type of action, and other external domain information, such as location is also recorded. Due to the company's secrecy policy, we will not name the other domains that is recorded associating with each event, but in total, information from over 20 domains are collected for the analysis. The following table is an example of the behavioral data collected in a tabular format.

\bigskip
\begin{center}
\captionof{table}{Example Data Format of the Multivariate Event Sequence}
 \begin{tabular}{| C{0.8cm}  C{1.2cm}  C{0.8cm}  C{1.1cm}  C{1.7cm}  C{1cm} |} 
 \hline
 Events Index & Page Type & Duration (sec) & Action Type & Location (lontitude, latitude) & Other Domains \\  
 \hline\hline
 1 & Register & 16 & submission &127.0223,31.2578 & \ldots  \\ 
 \hline
 2 & Login & 5 & submission & 127.0223,31.2578 & \ldots  \\
 \hline
 3 & ID Verification &  17 & get & 127.0223,31.2578 & \ldots  \\
 \hline
 4 & Loan Index  & 77 & click96 & 127.0223,31.2578 & \ldots \\
 \hline
 \vdots & \vdots & \vdots & \vdots &\vdots & $\ddots$  \\ [1ex] 
 \hline
\end{tabular}
\end{center}

As shown in the table, every application comes with a series of events that are components of categorical features from different domains. The data formats from both applications we discuss are very similar to the one we just described. 

\section{Session-based fraud detection in online e-commerce}

\begin{figure*}
\includegraphics[height=3in, width=6.5in]{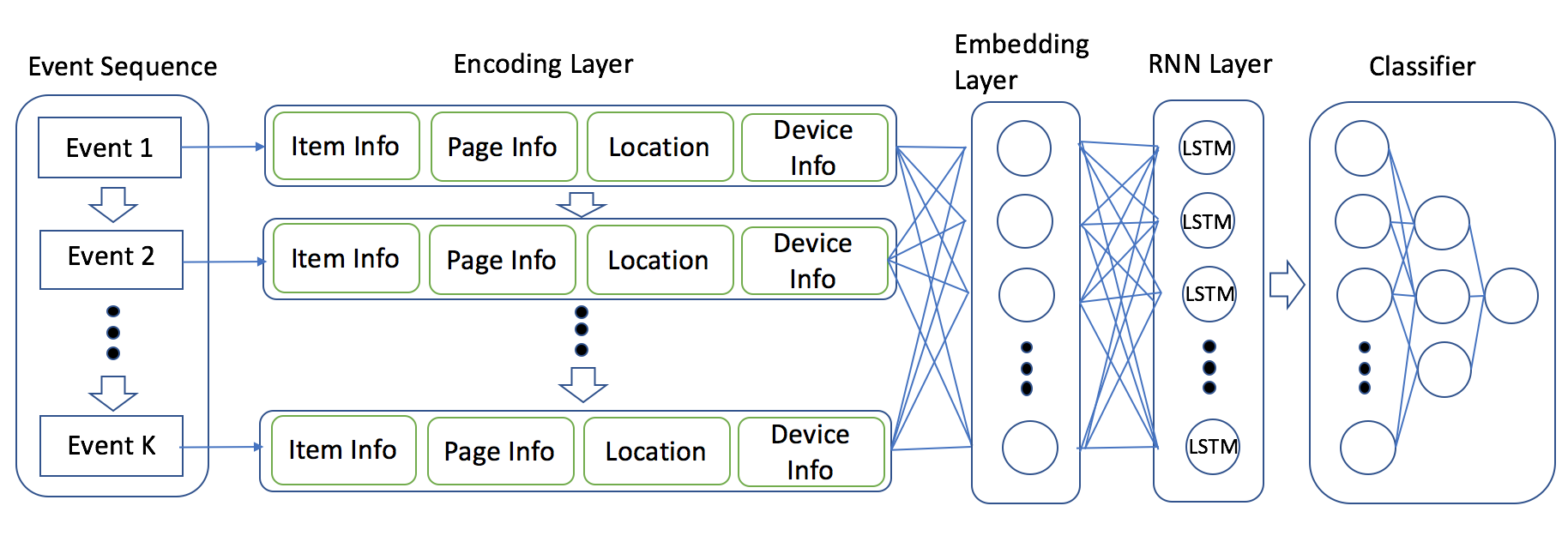}
\caption{Model Strucutre on Sequence of Events in E-commerce }
\end{figure*}

\subsection*{Business Background}

Two types of frauds exists in e-commerce platforms: account hijacking and fake bank card. Typically, fraudsters hack user's accounts to spend their account balance or use a fake card to register a new account. Traditional approaches using aggregated features, such as total spending amount or shopping times in the past month, were proven to be invalid or inapplicable since many instances of fraud are only detectable using their individual browsing actions in a website or application. We implement a RNN based Long Short Term Memory (LSTM) approach to capture the detailed user browsing behavior within a session, enabling the detection of fraud in our e-commerce platform.  Over time the LSTM cells learn to output, overwrite, or reset their internal memory based on their current input and history of past internal states, leading to a system capable of retaining information across hundreds of clicking actions within a session.

\subsection*{Network Structure}
RNN uses the idea of having cycles that feed the activation output from a previous time step as input to the network again to influence predictions at the current time step. All the actions stored in the internal states of the network theoretically should be held for long-term contextual information in a sequence; however, due to vanishing or exploiting gradients, past actions over 5 to 10 discrete time steps can be easily ``forgetten''  by the network. The LSTM solves this issue by adding a memory unit at every state with the idea of leveraging historical memory and current information for future predictions. In our practice, we found that by stacking multiple layers of LSTM for better learning of the sequence with high dimensionality on each state provided a better model performance.   \\

\medskip

Figure.1 demonstrates the underlying network structure we designed for processing the sequence in a session. Raw features in domains, such as the IP address, browser and its version, operation system, screen resolution, flash et al., are one-hot encoded into binary integers separately. Further, to process URL information, we convert it into five parts:a URL type, a category with three levels, and an item number. For the former two, we also use on-hot encoding. For the latter, due to the correlations among items, we adapt Word2Vec, regarding each item and session as a "word" and a "sentence", respectively. Thus, each item can be represented by a 32-dimension vector. Thus, the encoding layer encodes the session information associated with each clicking event into a 5204-dimension vector. Supposing the batch size is 128, a matrix M with a dimension of $128 \times200\times5204$ is fed to the RNN layer of the structure built by stacked the LSTM layers. The fully connected layers and a softmax group are followed.

\subsection*{Model Performance and Evaluation}
After 7000 epochs using Adagrad as the optimizer to train the neural network, the predicting accuracy exceeds our expectation on the testing dataset. Figure 2 shows the AUC and KS score of the model on the testing set. The precision and coverage rate to detect fraud on the testing set with 25,000 sessions with 790 of them being negatively labeled are 61\%  and 81\%, respectively. The AUC score is 0.96, and the Kolmogorov Smirnov score is 0.85. 

\begin{figure}[H]
\includegraphics[height=1.5in, width=1.5in]{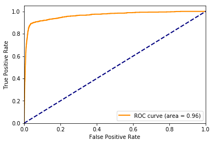}
\includegraphics[height=1.5in, width=1.5in]{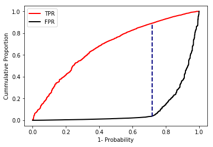}
\caption{AUC and KS diagram of Model on Testing Dataset}
\end{figure}

\section{Fraud Detection in Online Lending}

\begin{figure*}
\includegraphics[height=2.3in, width=6in]{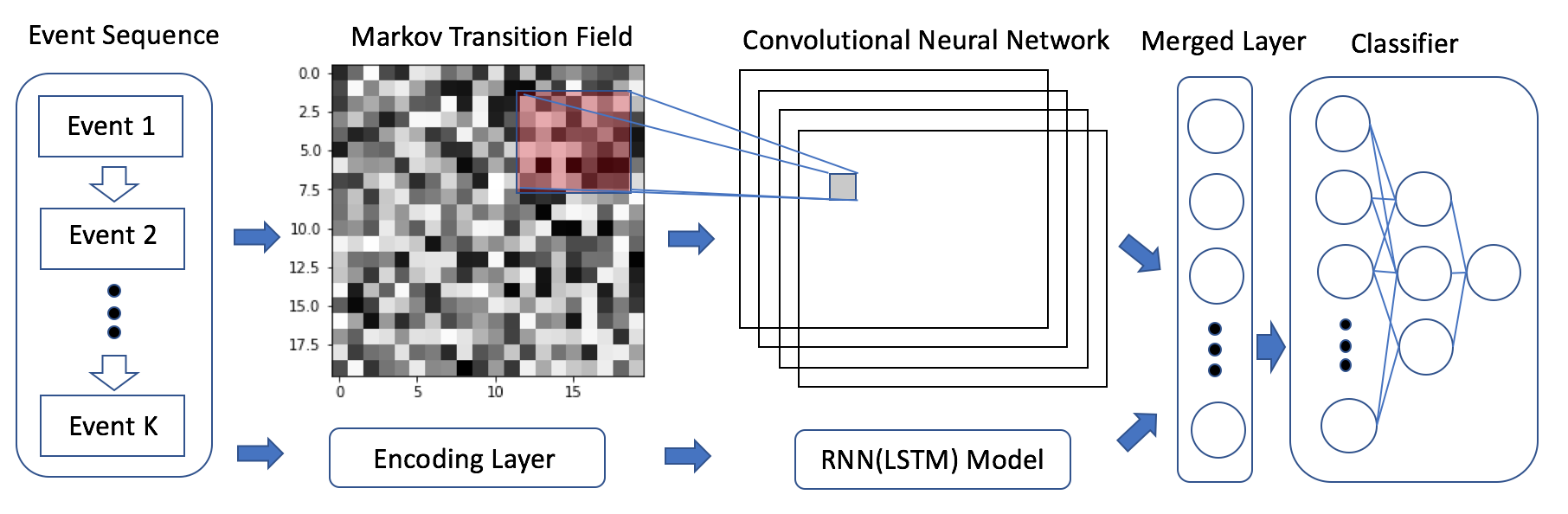}
\caption{Markov Transition Field Integrated Model}
\end{figure*}

\subsection*{Business Background}
In China, the online lending business has been one of the fastest growing financial services since 2016. According to the statistics \cite{statistics}, around 2000 online lending companies comprised over 200 billion RMB transactions in the first 10 months of 2017. Since this kind of financial service is for people with no guarantee nor mortgage, the sparse nature of credit-related features makes traditional models that use algorithms such as logistic regression ineffective towards potential fraudulent attacks. However, since many of these loan applications are submitted from either website or smartphone apps with instant needs for decision making, traditional approaches are incapable of extracting useful information for predicting from unstructured user behavioral data. With advanced data collection software implementations, financial smart-phone apps and website platforms were able to monitor and collect some of the user behaviors with the users' permissions. To replace traditional models built on credit-related features, we explored applying deep learning techniques and MTF to build a model framework capable of processing users' behavioral interactions with smartphone apps and accurately estimating potential fraud risks.

\subsection*{Integrating Markov Transition as Embedding to Convolutional Network}
We propose a framework similar to Wang and Oate's work \cite{Wang Oates} for encoding dynamic transitional fields in a sequence of actions by using the idea of preserving the information in the inter-relationships of activities by extracting the Markov transition probabilities. After generating a 2-D MTF $M$ from a sequence, we let a convolutional network learn information from $M$ similar to image networks, such as ImageNet or VGG16. Since each element in $M$ is calculated as a transition probability, for the choice of pooling layer, average pooling layers are applied instead of max-pooling layers to avoid loss of information. However, the information contained within the order of the actions in a sequence is lost in this method of encoding because MTF only considers the likelihood of an action transitioning to the next. To enhance the final predictive accuracy, an RNN model similar to the one discussed in the previous section is added. The final classifier takes outputs from a convolutional network on the extracted MTF $M$ and a recurrent network on encoded sequences. 

\bigskip

Given a series of actions, first, similar to the preprocessing stage in the previous section, suppose for every action $a_i$ there is a total number of $k$ features. Each action $a_i$  is first encoded by a method similar to a label encoder:  all features are bin and one-hot coded into a binary vector. The indexes of the binary vector are then taken with a maximum numerical value of the length ($l$) of the binary vectors, denoted as $V_i$. Each feature $f_i$ is coded with a value of 1 on $V_{ij}$, where $ a \leqslant j < b$ on its special segment $s_i = [a,b)$. In this way, every vector $V_i$ is guaranteed $k$ non-zero numerical values. A Markov Transition Field $M$ is constructed by taken all non-zero numerical values from $V_i$ to $V_{i+1}$. The meaning of a transition probability $m_{ij}$ in $M$ can be interpreted as the probability of the likelihood of transition from $f_x$ to $f_y$ where $i \in s_x$ and $j \in s_y $:

\begin{figure}[H]
\includegraphics[height=1.4in, width=3in]{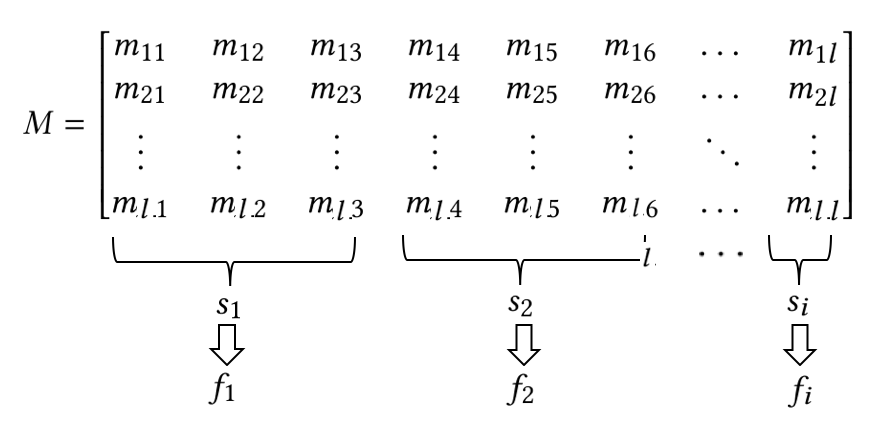}
\end{figure}
 
 An MTF is constructed by a sequence of actions for each user with the dimension of $l \times l $. Suppose $n$ records are feed to the network in every batch, similar to the convolutional network for image processing, and a matrix with dimension of $ n \cdot l \cdot l$ is fed to the network. 

\subsection*{Model Performance and Evaluation}
In general, the network stucture model is able to outperform the traditional k-nearest-neighbor classifier based on DTW as distance metric in our experiment. By combining the RNN model and a CNN built on MTF, the combined model outperformed a simple multilayer perception classifier by 23\% and a single RNN model by 7\%. Kolmogorov Smirnov scores increased from 0.17 in baseline model to 0.21 in the integrated MTF deep learning model. Considering the Kolmogorov Smirnov score for a general pre-loan predicting model on a single domain features is typically under 0.20, the result of this model is considered very good. The model combines the advantage of having a representing information in a sequence from a general picture using the MTF learned by a convolutional network and information in order of states in sequence learned by the RNN. 

\begin{figure}[H]
\includegraphics[height=0.9in, width=3in]{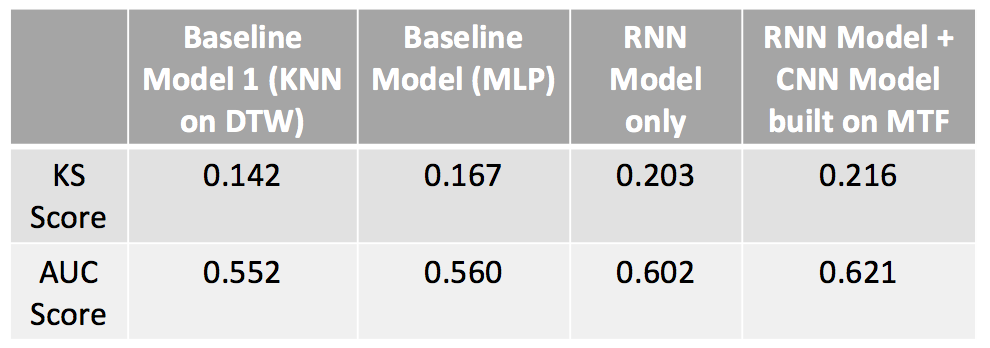}
\caption{Table of Performance Metric on Baseline Model and Proposed Model}
\end{figure}

\section{Conclusion}
In this paper, we discussed some of the challenges of using weak state-streaming data in online fraud detection in online shopping and finance scenarios. We proposed two neural network structures, one built based on encoding the sequence and learning it using the RNN constructed by LSTM cells, and the other built by using the CNN learned upon the Markov-Transition Field extracted from the sequence. Both modeling structures quantify and format the state-streaming data into a workable structured format with a classifier for prediction. We tested both models on real massive datasets and concluded that RNN built by LSTM nodes is effective in processing sequential behavioral data and integrating the data with a CNN with the encoded MTF can boost model performance.

\bibliographystyle{ACM-Reference-Format}
{}

\end{document}